\documentclass[10pt]{article} % 建议使用双栏，更具学术感
\usepackage[utf8]{inputenc}
\usepackage[margin=0.75in]{geometry} % 调整边距
\usepackage{authblk} % 专门处理作者栏的宏包
\usepackage[numbers,sort&compress]{natbib} % 推荐用 natbib，比 cite 更强大
\usepackage{url} 
\usepackage{amsmath, amssymb, amsfonts} % 处理数学公式和 \text 命令
\usepackage{graphicx}                   % 处理插图和 \includegraphics
\title{\textbf{InfiAgent: An Infinite-Horizon Framework for General-Purpose Autonomous Agents}}
\usepackage{booktabs}

\author[1]{\textbf{Chenglin Yu}}
\author[2]{\textbf{Yuchen Wang}}
\author[2]{\textbf{Songmiao Wang}}
\author[2]{\textbf{Hongxia Yang}}
\author[2]{\textbf{Ming Li}\thanks{Corresponding author: ming.li@polyu.edu.hk}}

\affil[1]{The University of Hong Kong}
\affil[2]{The Hong Kong Polytechnic University}

\date{} % 隐藏日期

\begin{document}
\maketitle

\begin{abstract}
LLM agents can reason and use tools, but they often break down on long-horizon tasks due to unbounded context growth and accumulated errors. Common remedies such as context compression or retrieval-augmented prompting introduce trade-offs between information fidelity and reasoning stability. We present \textbf{InfiAgent}, a general-purpose framework that keeps the agent’s reasoning context strictly bounded regardless of task duration by externalizing persistent state into a \emph{file-centric state abstraction}. At each step, the agent reconstructs context from a workspace state snapshot plus a fixed window of recent actions. Experiments on DeepResearch and an 80-paper literature review task show that, without task-specific fine-tuning, InfiAgent with a 20B open-source model is competitive with larger proprietary systems and maintains substantially higher long-horizon coverage than context-centric baselines. These results support explicit state externalization as a practical foundation for stable long-horizon agents.

Repo:https://github.com/ChenglinPoly/infiAgent
\end{abstract}


\section{Introduction}

Large Language Models (LLMs) have increasingly been deployed as autonomous agents capable of planning, tool use, and multi-step reasoning across a wide range of tasks~\cite{naveed2025comprehensive, tran2025multi}. Recent agent-based systems demonstrate promising performance in domains such as scientific research, software engineering, and information synthesis, suggesting that LLMs can serve as general-purpose problem solvers that coordinate external tools and resources. Despite this progress, existing LLM agents remain brittle when deployed over long task horizons.

A core challenge arises from how agent state is represented and maintained. Most current agent frameworks implicitly treat the LLM prompt as the primary carrier of state, accumulating dialogue history, tool traces, intermediate plans, and partial results directly within the context window. As task duration increases, this design leads to unbounded context growth, forcing systems to rely on truncation, summarization, or heuristic retrieval to remain within model limits. These mechanisms introduce well-known failure modes, including information loss, interference from irrelevant tokens, and increased sensitivity to early errors, ultimately resulting in unstable behavior over long horizons.

Retrieval-Augmented Generation (RAG) and long-context models partially mitigate context length limitations by enabling access to external memory or larger windows. However, these approaches still entangle long-term task state with the agent’s immediate reasoning context, placing increasing cognitive load on the LLM as execution progresses. Empirical evidence suggests that agent performance often degrades as context fills, a phenomenon sometimes described as the “illusion of state” in autonomous research agents~\cite{shojaee2025illusion}. Consequently, simply extending context length does not fundamentally resolve the long-horizon stability problem.

Recent work such as MAKER~\cite{meyerson2025maker} demonstrates that near-infinite execution is possible in highly structured domains through extreme task decomposition. While effective for logic-based problems with predefined subtask boundaries, such approaches rely on specialized micro-agents and rigid workflows, limiting their applicability to open-ended domains like scientific research, where task structure, intermediate goals, and relevant information sources are not known a priori.

In this work, we argue that achieving stable long-horizon behavior in LLM agents requires an explicit separation between \emph{persistent task state} and \emph{bounded reasoning context}. We introduce \textbf{InfiAgent}, a general-purpose agent framework that externalizes long-term state into a file-centric representation. Rather than storing historical information implicitly within the prompt, InfiAgent treats the file system as the authoritative and persistent record of the agent’s actions, environment, and intermediate artifacts. At each decision step, the agent reconstructs its reasoning context solely from a snapshot of this externalized state and a fixed-size window of recent actions, ensuring that the context size remains strictly bounded regardless of task duration.

Building on this state abstraction, InfiAgent employs a hierarchical agent architecture that enforces structured task decomposition and controlled tool invocation. High-level planning agents operate over abstract goals and state summaries, while lower-level agents execute domain-specific or atomic actions. This design reduces error propagation commonly observed in flat multi-agent systems and enables consistent behavior across extended execution horizons. In addition, InfiAgent introduces an external attention mechanism that processes large documents and heavy information outside the main reasoning context, injecting only task-relevant outputs back into the agent’s state.

We evaluate InfiAgent on the DeepResearch benchmark~\cite{shojaee2025illusion} and a long-horizon literature review task involving up to 80 academic papers. Without task-specific fine-tuning, InfiAgent equipped with a 20B-parameter open-source model achieves performance comparable to or exceeding larger proprietary agents on DeepResearch. In extended evaluations spanning hundreds of execution steps, the proposed file-centric state abstraction enables reliable task completion, whereas baseline agents frequently degrade due to context limitations. These results suggest that explicit state externalization provides a practical and effective foundation for building stable, general-purpose LLM agents capable of operating over arbitrarily long horizons.

\section{Related Work}

\subsection{Multi-Agent System Architectures}
Traditional multi-agent systems have primarily focused on peer-to-peer collaboration models. While effective for simple coordination, these approaches struggle with complex, hierarchical task decomposition. Recent work has explored hierarchical organization~\cite{hierarchical2025}. Moore~\cite{taxonomy2025} categorized Hierarchical Multi-Agent Systems (HMAS) highlighting trade-offs between global efficiency and local autonomy. Comprehensive reviews note a persistent challenge in designing unified agents that seamlessly integrate cognition, planning, and interaction~\cite{agentReview2025}. InfiAgent addresses these by enforcing stability through strict compositional constraints and a file-centric state.

\subsection{Task Decomposition in Multi-Agent Systems}
Task decomposition has been a central challenge. Existing approaches include goal-oriented~\cite{goal2023}, constraint-based~\cite{constraint2017}, and learning-based decomposition~\cite{learning2022}. Frameworks like AGENTiGraph leverage knowledge graphs (KGs) for decomposition~\cite{KGAgent2025}. However, these approaches often lack guarantees regarding system stability over long horizons. MAKER~\cite{meyerson2025maker} introduced extreme decomposition for logic tasks, but its applicability to open-ended research remains limited. InfiAgent employs a recursive DAG-based decomposition that is flexible enough for general tasks while maintaining strict parent-child control.

\subsection{Long-Context and Autonomous Research Agents}
The "illusion of state" in deep research agents has been highlighted by Shojaee et al.~\cite{shojaee2025illusion}, showing that performance often degrades as context fills. Agents like AgentGym~\cite{xi2024agentgym} and Richelieu~\cite{guan2024richelieu} focus on self-evolution, while systems like the AI Scientist attempt end-to-end automation. However, many rely on extended context windows or manual templates. InfiAgent's approach of "Zero Context Compression" via file-based state offers a robust alternative to these context-heavy methods.

\section{Formalizing File-Centric State for Long-Horizon Agents}
\label{sec:formalization}

We formalize the long-horizon agent execution problem by distinguishing between \emph{persistent task state} and \emph{bounded reasoning context}. This separation clarifies the limitations of context-centric agent designs and motivates the file-centric state abstraction adopted by InfiAgent.

\subsection{Agent Execution as a State-Conditioned Decision Process}

We consider an autonomous LLM agent operating over discrete time steps $t = 1, 2, \dots$. At each step, the agent selects an action $a_t \in \mathcal{A}$ conditioned on an internal representation of task state. In conventional agent frameworks, this state is implicitly represented by the accumulated prompt context:
\begin{equation}
    c_t = \langle o_1, a_1, \dots, o_{t-1}, a_{t-1}, o_t \rangle,
\end{equation}
where $o_t$ denotes observations such as user instructions, tool outputs, or intermediate results. As $t$ increases, the context length $|c_t|$ grows unbounded, necessitating truncation or compression to fit within a finite context window.

We define this design as \emph{context-centric state representation}, where long-term task information and short-term reasoning signals are entangled within a single sequence. This entanglement introduces an inherent trade-off between information retention and reasoning stability, particularly in long-horizon tasks.

\subsection{Persistent State Externalization}

To decouple long-term memory from bounded reasoning context, we introduce an explicit \emph{persistent state} representation $S_t$ that evolves over time:
\begin{equation}
    S_t = \mathcal{F}_t,
\end{equation}
where $\mathcal{F}_t$ denotes the set of files and structured artifacts stored in the agent's workspace at step $t$. These artifacts may include intermediate results, plans, summaries, datasets, or generated code, and collectively serve as the authoritative record of task progress.

The persistent state evolves through state-transition operators induced by agent actions:
\begin{equation}
    \mathcal{F}_{t+1} = \mathcal{T}(\mathcal{F}_t, a_t),
\end{equation}
where $\mathcal{T}$ captures file creation, modification, or deletion. Importantly, $\mathcal{F}_t$ is not subject to context window constraints and can grow with task complexity and duration.

\subsection{Bounded Reasoning Context Reconstruction}

At each step $t$, the agent constructs a bounded reasoning context $c_t^{\text{bounded}}$ by querying the persistent state:
\begin{equation}
    c_t^{\text{bounded}} = g(\mathcal{F}_t, a_{t-k:t-1}),
\end{equation}
where $a_{t-k:t-1}$ denotes a fixed-size window of the most recent $k$ actions, and $g(\cdot)$ is a deterministic context-construction function. In InfiAgent, $k$ is a small constant (e.g., $k=10$), ensuring that $|c_t^{\text{bounded}}| = \mathcal{O}(1)$ with respect to the task horizon.

This formulation guarantees that the agent’s reasoning context remains strictly bounded for all $t$, while full task history is preserved implicitly through the persistent state $\mathcal{F}_t$. Unlike summarization-based approaches, no information is discarded from the authoritative state; instead, relevance is determined dynamically at each step via state inspection.

\subsection{Comparison and Implications}

Compared to context-centric agents, which encode all task state directly in the prompt, and retrieval-augmented agents, which partially externalize memory but re-inject retrieved text into the context, the proposed file-centric abstraction treats persistent state as a first-class object. By reconstructing reasoning context through a bounded interface, InfiAgent eliminates unbounded context growth and reduces interference between long-term artifacts and short-term decision-making.

As a result, reasoning errors are less likely to accumulate implicitly over time, and the agent can scale execution horizon without increasing cognitive load on the underlying language model.

\section{Methodology: The InfiAgent Framework}
\label{sec:method}

\begin{figure*}[t]
    \centering
    \includegraphics[width=\linewidth]{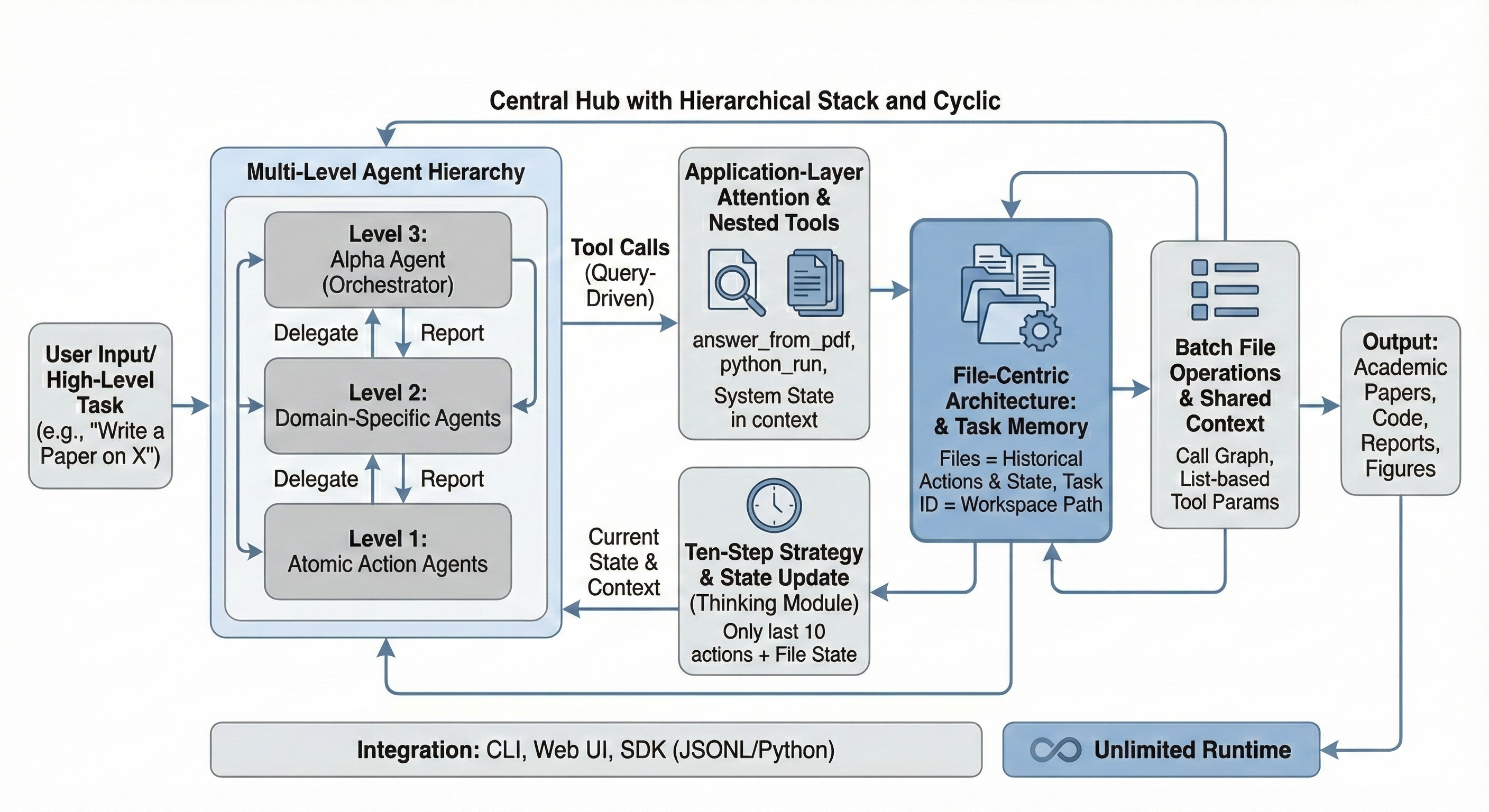}
    \caption{\textbf{The InfiAgent Framework.} InfiAgent implements a hierarchical execution stack over a file-centric persistent state. Files serve as the authoritative task memory, while an external attention mechanism processes heavy documents outside the bounded reasoning context. Periodic state consolidation refreshes the agent’s context from the workspace snapshot.}
    \label{fig:framework}
\end{figure*}

Building on the file-centric state formulation in Section~\ref{sec:formalization}, we describe InfiAgent as a concrete system that instantiates persistent state externalization and bounded context reconstruction. This section focuses on architectural and implementation choices, rather than re-defining the state abstraction.
Figure~\ref{fig:framework} provides an overview.

\subsection{File-Centric State Management and Task Memory}
\label{sec:file_centric}

InfiAgent materializes the persistent state $\mathcal{F}_t$ as a structured workspace on the file system. Each task is assigned a dedicated workspace directory that stores plans, intermediate artifacts, tool outputs, and validation logs. This workspace serves as the authoritative record of task progress and persists across execution steps and sessions.

To support bounded reasoning, InfiAgent maintains a fixed-length buffer of recent actions, which is combined with a snapshot of the workspace to construct the agent’s reasoning context at each step. In practice, this buffer is small (e.g., 10 actions), preserving short-term execution coherence without increasing context size.

\paragraph{Periodic State Consolidation.}
To sustain long-horizon execution, InfiAgent periodically consolidates recent progress into persistent state. At fixed intervals, high-level plans and progress markers stored in the workspace are updated, after which the reasoning context is refreshed using the updated state snapshot. This mechanism realizes bounded context reconstruction in practice while preserving detailed task artifacts in persistent storage.

\subsection{Multi-Level Agent Hierarchy}
InfiAgent organizes agents into a tree-structured hierarchy (DAG):
\begin{itemize}
    \item \textbf{Level 3 (Alpha Agent):} The orchestrator responsible for high-level planning and decomposing user requests into subtasks. It acts as the root of the decision tree.
    \item \textbf{Level 2 (Domain Agents):} Specialists such as the \texttt{Coder Agent}, \texttt{Data Collection Agent}, or \texttt{Paper Writer}. They execute specific workflows delegated by the Alpha Agent.
    \item \textbf{Level 1 (Atomic Agents):} Agents that handle specific tool executions, such as web searching or file I/O.
\end{itemize}
This hierarchy allows for serial execution with clear boundaries. Higher-level agents invoke lower-level agents as callable tools (Agent-as-a-Tool), preventing "tool calling chaos" often seen in flat multi-agent systems where agents compete for execution.

\subsection{External Attention Pipeline}
To handle massive information (e.g., reading 80 papers) without context bloat, InfiAgent uses an \textit{External Attention Pipeline}.
When an agent needs information from a document, it does not load the document into its context. Instead, it calls a specialized tool (e.g., \texttt{answer\_from\_pdf}). This tool spins up a temporary, isolated LLM process to query the document and returns only the extracted answer.
\begin{equation}
    C_{main} \leftarrow C_{main} \cup Tool(Query, Document)
\end{equation}
This effectively offloads the "reading" cost to the tool layer, keeping the main agent's cognitive load low and focused on decision-making. This mechanism acts as an application-layer attention head, selecting only the relevant information from massive external data sources.

\section{Experiments}

The goal of our experiments is to evaluate whether explicit state externalization improves the stability and reliability of LLM agents in long-horizon tasks. Rather than optimizing for peak performance on short interactions, we focus on execution robustness under extended task duration, large document collections, and repeated tool usage.

We evaluate InfiAgent along two complementary dimensions. First, we assess general research capability using the DeepResearch benchmark~\cite{shojaee2025illusion}, a standardized evaluation designed to measure multi-step research quality. Second, we design a long-horizon literature review task to directly stress-test agent stability over hundreds of execution steps. Across all experiments, we use identical model backbones and comparable prompting budgets when applicable, and we report averaged results over multiple runs to reduce variance.

\begin{figure*}[t]
    \centering
    \includegraphics[width=\linewidth]{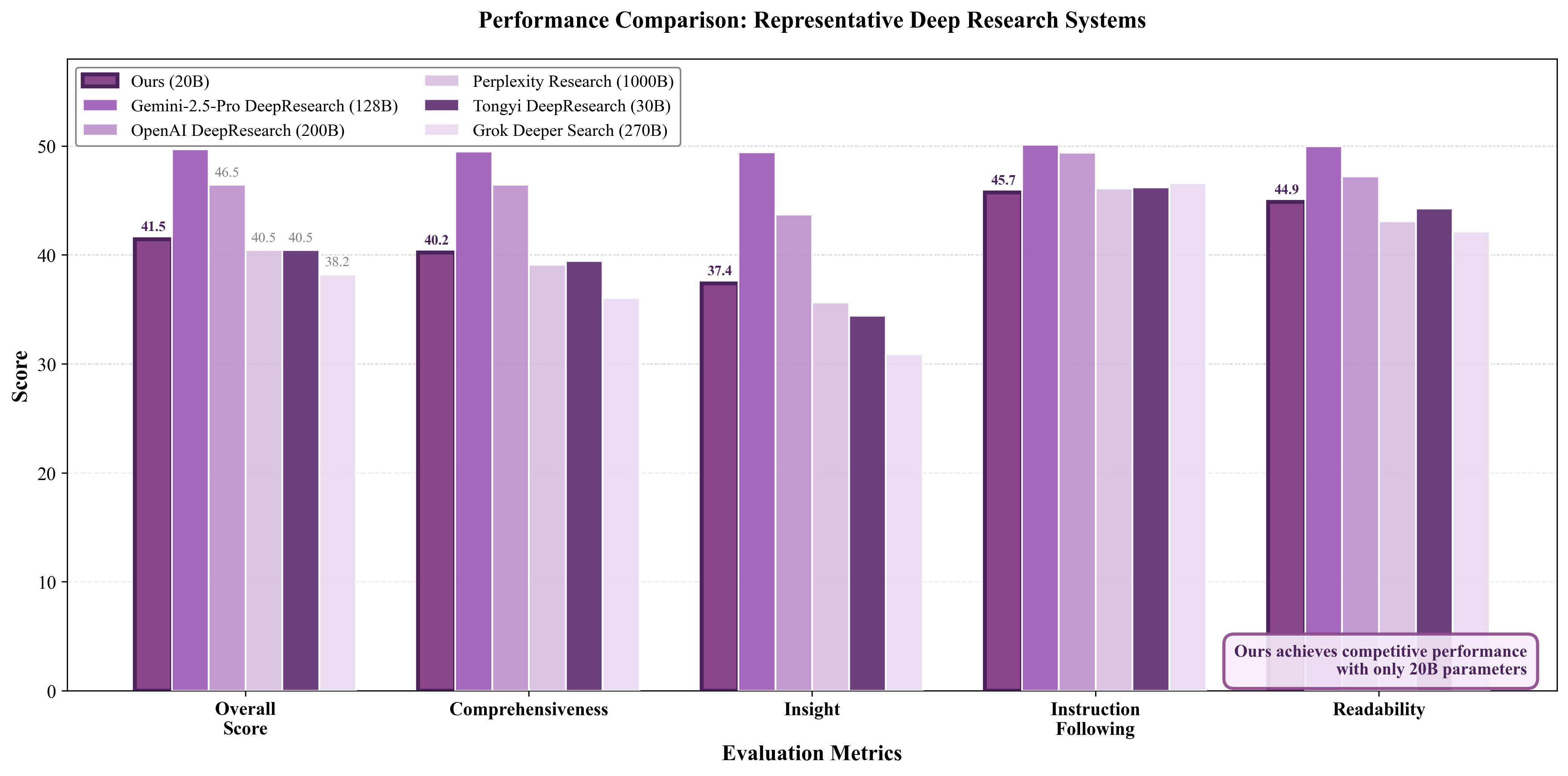}
    \caption{\textbf{Component-wise comparison on DeepResearch.} Scores are broken down by evaluation dimension. InfiAgent shows strong performance on instruction following and readability, which are closely related to structured state management and output control.}
    \label{fig:comparison}
\end{figure*}

\subsection{DeepResearch Benchmark}
\label{sec:deepresearch}

We evaluate InfiAgent on the DeepResearch benchmark~\cite{shojaee2025illusion}, which is designed to assess an agent’s ability to conduct multi-step research involving information gathering, synthesis, and structured reporting. The benchmark scores agents along four dimensions: comprehensiveness, insight, instruction following, and readability.

\paragraph{Setup.}
We configure InfiAgent using a 20B-parameter open-source model (\texttt{gpt-oss-20b}) without any task-specific fine-tuning. To isolate the impact of agent architecture, we follow the standard benchmark protocol and adopt default task instructions wherever possible. All reported scores are taken directly from the benchmark evaluation to ensure comparability across systems.

\begin{figure}[t]
    \centering
    \includegraphics[width=\linewidth]{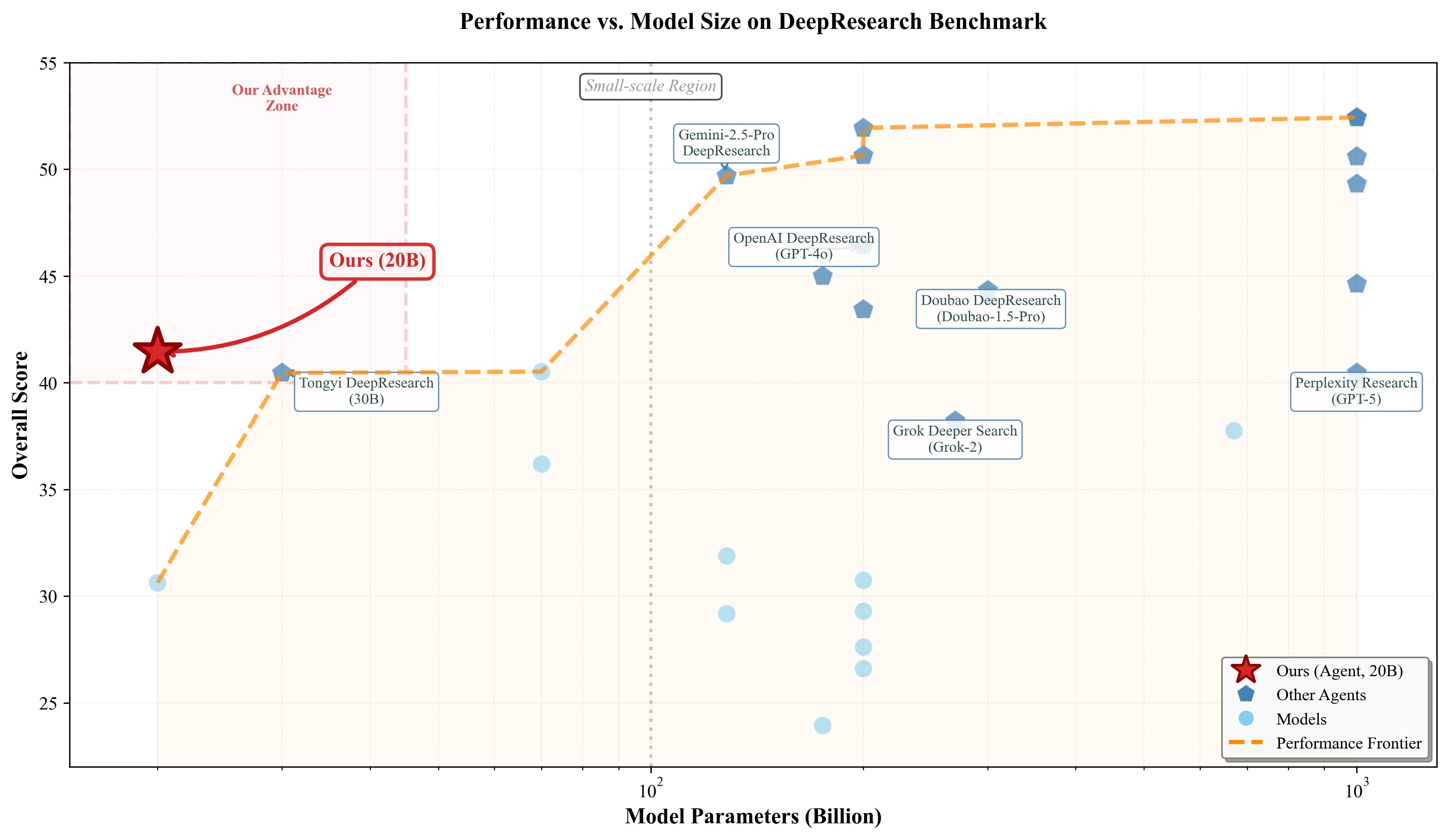}
    \caption{\textbf{Performance vs. model size on DeepResearch.} InfiAgent (20B) achieves competitive performance relative to larger proprietary agents evaluated on the same benchmark, suggesting an improved efficiency–performance trade-off.}
    \label{fig:performance_frontier}
\end{figure}

\paragraph{Overall performance.}
Figure~\ref{fig:performance_frontier} plots DeepResearch scores against model size. InfiAgent achieves an overall score of 41.45 with a 20B model, placing it on a favorable efficiency frontier relative to systems that rely on substantially larger models. While absolute performance varies across agents, this result indicates that architectural design can partially compensate for differences in model scale in long-form research tasks.

\paragraph{Component-wise analysis.}
Figure~\ref{fig:comparison} presents a breakdown across evaluation dimensions. InfiAgent performs particularly well on instruction following and readability. We attribute this behavior to the explicit file-centric state and structured execution pipeline, which encourage adherence to task instructions and consistent output formats. Performance on insight and comprehensiveness is competitive with larger models, suggesting that stable state management supports sustained reasoning over extended interactions.

Overall, these results suggest that explicit externalization of agent state can improve the reliability and efficiency of multi-step research behavior, even when using smaller backbone models.

\subsection{Long-Term Literature Review Task}
\label{sec:long_term_task}

To evaluate long-horizon stability, we design a literature review task that requires an agent to iterate over a large paper collection. Each run provides a list of 80 academic papers and instructs the agent to (i) read each paper, (ii) produce a short summary, and (iii) assign a relevance score. This task stresses sustained tool usage and state tracking over hundreds of steps.

\paragraph{Evaluation metric (coverage).}
We focus on \emph{task completion reliability} rather than summary quality. Specifically, we measure \textbf{coverage}, defined as the number of papers for which the agent produces a non-empty summary that is grounded in the paper content (i.e., the summary contains at least one content-based statement beyond title/metadata). We report the maximum, minimum, and average coverage over repeated runs under the same setting.

\begin{table}[h]
    \centering
    \resizebox{\columnwidth}{!}{
    \begin{tabular}{llccc}
        \toprule
        \textbf{Setting} & \textbf{Model} & \textbf{Max} & \textbf{Min} & \textbf{Avg} \\
        \midrule
        \multicolumn{5}{l}{\textbf{Main results (with file-centric state; InfiAgent vs. baselines)}} \\
        \midrule
        \textbf{InfiAgent} & GPT-OSS-20B & \textbf{80} & 15 & 67.1 \\
        \textbf{InfiAgent} & Gemini-3-Flash & \textbf{80} & \textbf{80} & \textbf{80.0} \\
        \textbf{InfiAgent} & Claude-4.5-Sonnet & \textbf{80} & \textbf{80} & \textbf{80.0} \\
        Claude Code & Claude-4.5-Sonnet & 80 & 11 & 29.1 \\
        Cursor & Claude-4.5-Sonnet & 5 & 0 & 1.0 \\
        Cursor & Gemini-3-Flash & 1 & 0 & 0.1 \\
        \midrule
        \multicolumn{5}{l}{\textbf{Ablation (remove file-centric state; compressed long-context prompts)}} \\
        \midrule
        No File State (Compressed Context) & GPT-OSS-20B & 7 & 1 & 3.2 \\
        No File State (Compressed Context) & Gemini-3-Flash & 25 & 20 & 21.1 \\
        No File State (Compressed Context) & Claude-4.5-Sonnet & 77 & 11 & 27.7 \\
        \bottomrule
    \end{tabular}
    }
    \caption{\textbf{Long-horizon task reliability (coverage) and ablation.} Coverage on the 80-paper literature review task. Top: main results with file-centric state (InfiAgent vs. baselines). Bottom: ablation removing file-centric state and replacing it with compressed long-context prompts. We report max/min/avg coverage across repeated runs.}
    \label{tab:long_term}
\end{table}

\paragraph{Results.}
Table~\ref{tab:long_term} shows that InfiAgent consistently achieves high coverage, processing the full set of 80 papers with stronger backbones and maintaining substantial coverage even with a 20B model. This behavior is consistent with the proposed file-centric state abstraction, which enables the agent to iterate through long document lists without relying on an ever-growing prompt history.

In contrast, baseline agents exhibit substantially lower average coverage and high variance across runs. Manual inspection suggests that failures often manifest as early termination, skipping items, or producing summaries that primarily restate titles rather than paper content.

\paragraph{Ablation analysis.}
Removing the file-centric state and relying on compressed long-context prompts substantially degrades coverage across all models (bottom block of Table~\ref{tab:long_term}). Even with stronger backbones, the ablated variant shows reduced average coverage and large run-to-run variance. This supports the claim that explicit persistent state externalization is a key contributor to long-horizon stability and cannot be fully replaced by long-context compression alone.

\paragraph{Limitations of the comparison.}
We do not claim absolute superiority over proprietary systems. Commercial agents may employ undocumented optimizations or termination heuristics that trade off completeness for latency or cost. Our goal is to highlight a qualitative difference in long-horizon behavior: explicit externalization of persistent state is associated with higher and more stable coverage under extended execution horizons.

\subsection{Case Study: Real-World Applications and Expert Blind Review}

To validate the framework's capability, we implemented \textbf{InfiHelper}, a semi-general agent powered by the InfiAgent framework. While InfiAgent provides the underlying architecture (File-Centric State, Hierarchy, etc.), InfiHelper is the concrete instantiation designed for complex knowledge work.

\paragraph{Versatile Application Scenarios of InfiHelper}
InfiHelper is not limited to a single domain; it functions as a versatile assistant capable of handling diverse real-world tasks using the same core architecture:
\begin{itemize}
    \item \textbf{Computational Biology}: InfiHelper successfully conducted "dry-lab" experiments, such as simulating Extra-Cellular Matrix (ECM) protein compositions and predicting interactions.
    \item \textbf{Logistics \& Operations}: The same agent performed automated shift scheduling for logistics companies, optimizing workforce allocation under complex constraints.
    \item \textbf{Academic Research}: InfiHelper functioned as an automated research assistant capable of end-to-end scientific discovery, from literature review to manuscript drafting.
\end{itemize}

\paragraph{Expert Blind Review on Academic Output}
To assess the quality of complex outputs, we focused on the academic research scenario. We tasked InfiHelper with generating full-length research papers and submitted them to a blind review panel of experts with EI/IEEE conference experience.
The results indicated that the papers generated by InfiHelper achieved \textbf{Human-Level Quality}, with reviewers judging them as satisfying the acceptance criteria for standard academic conferences. The reviewers highlighted the logical coherence of the experiments and the correctness of the \LaTeX{} formatting, attributes directly supported by the underlying InfiAgent framework's stable state management and rigid output constraints. This demonstrates that InfiHelper, enabled by InfiAgent, can effectively support, and in some cases automate, rigorous knowledge work across different domains.

\section{Discussion}

\paragraph{What Explicit State Externalization Does Not Solve.}
While the proposed file-centric state abstraction significantly improves long-horizon stability, it does not inherently enhance the reasoning capability of the underlying language model. If a backbone model produces incorrect intermediate conclusions or flawed artifacts, these errors may still be written into persistent state and propagated across subsequent steps. Although separating persistent state from bounded context enables inspection and potential correction, it does not eliminate the need for validation mechanisms or human oversight in high-stakes settings.

\paragraph{Long Context Is Not a Substitute for Persistent State.}
The ablation results in Section~\ref{sec:long_term_task} highlight an important distinction between long context and persistent state. Even when using models with large context windows, replacing file-centric state with compressed long-context prompts leads to substantial degradation in coverage and increased variance across runs. This suggests that increased context length alone is insufficient for reliable long-horizon execution, and that explicit state externalization plays a qualitatively different role than context expansion.

\paragraph{Efficiency and Latency Trade-offs.}
Externalizing state and enforcing hierarchical execution introduce additional overhead compared to single-pass, context-centric agents. File system operations, periodic state consolidation, and serialized agent execution may increase latency, particularly for tasks requiring rapid responses. As a result, the proposed framework is better suited for long-running, knowledge-intensive workflows than for real-time interactive applications. Exploring asynchronous updates and partial parallelism remains an important direction for future work.

\paragraph{Scope and Generalization.}
Our empirical evaluation focuses on research-oriented tasks involving extended document processing and multi-step information synthesis. While these settings expose long-horizon failure modes, they do not fully represent other agent scenarios such as reactive dialogue, embodied interaction, or environments with rapidly changing external state. Further evaluation across a broader range of tasks is needed to assess generality.

\paragraph{Broader Implications.}
Despite these limitations, the results suggest a general design principle for long-horizon agents: persistent task state should be treated as a first-class object, distinct from the bounded reasoning context of the language model. This separation enables systematic analysis of agent behavior over time and provides a foundation for future work on verification, correction, and collaborative multi-agent systems.

\section{Limitations}
While InfiAgent demonstrates robust capabilities in long-horizon tasks, several limitations remain.
First, the multi-level agent hierarchy introduces \textbf{latency overhead}. As the depth of the agent tree increases, the time required for task delegation and result aggregation grows, potentially impacting real-time responsiveness.
Second, despite the file-centric state management mitigating context loss, \textbf{hallucination accumulation} is still a risk, particularly when using smaller models (e.g., 20B) for extremely long tasks. If a sub-agent writes incorrect information to the file system that is not caught by the validation mechanism, downstream agents may propagate this error.
Third, the current architecture strictly enforces serial execution to ensure state consistency, which means it \textbf{does not support parallel processing}. This limits the system's efficiency in tasks that are inherently parallelizable, such as simultaneous literature review and experiment coding.

\section{Conclusion}

InfiAgent proposes a paradigm shift from context-based agents to file-centric agents. By externalizing state into the file system and employing a rigorous multi-level hierarchy, InfiAgent achieves effectively infinite runtime and high stability. Our results demonstrate that this \textbf{training-free} architecture enables smaller open-source models (20B) to compete with state-of-the-art proprietary agents on complex research benchmarks, paving the way for more accessible and scalable autonomous research systems.

\section*{AI Usage Declaration}
In accordance with ACL policies, we declare the use of AI assistance in the preparation of this manuscript. The framework illustration (Figure \ref{fig:framework}) was generated using the Gemini Nano Banana model. The text of this paper was polished for clarity and grammar using Gemini 1.5 Flash. All scientific claims, experimental designs, and results were verified by human authors.

\bibliographystyle{unsrtnat}
\bibliography{reference}

\appendix

\section{Appendix: Detailed Experimental Results}
\label{sec:appendix}

Table \ref{tab:detailed_results} provides the complete evaluation results for InfiAgent and various baseline systems on the DeepResearch benchmark.

\begin{table*}[h]
    \centering
    \small
    \resizebox{\linewidth}{!}{
    \begin{tabular}{lcccccccc}
        \toprule
        \textbf{Model/Agent} & \textbf{Overall} & \textbf{Comp.} & \textbf{Insight} & \textbf{Inst. Fol.} & \textbf{Read.} & \textbf{Cit. Acc.} & \textbf{Eff. Cit.} & \textbf{Params (B)} \\
        \midrule
        tavily-research (GPT-5) & 52.44 & 52.84 & 53.59 & 51.92 & 49.21 & - & - & 1000 \\
        thinkdepthai-deepresearch (GPT-5) & 52.43 & 52.02 & 53.88 & 52.04 & 50.12 & - & - & 1000 \\
        cellcog (GPT-4o) & 51.94 & 52.17 & 51.90 & 51.37 & 51.94 & - & - & 200 \\
        salesforce-air-deep-research (GPT-4o) & 50.65 & 50.00 & 51.09 & 50.77 & 50.32 & - & - & 200 \\
        gensee-search-gpt-5 (GPT-5) & 50.60 & 50.06 & 50.76 & 51.31 & 49.72 & 32.94 & 21.06 & 1000 \\
        gemini-2.5-pro-deepresearch & 49.71 & 49.51 & 49.45 & 50.12 & 50.00 & 78.30 & 165.34 & 128 \\
        langchain-open-deep-research (GPT-5) & 49.33 & 49.80 & 47.34 & 51.05 & 48.99 & 34.74 & 22.44 & 1000 \\
        openai-deepresearch (GPT-4o) & 46.45 & 46.46 & 43.73 & 49.39 & 47.22 & 75.01 & 39.79 & 200 \\
        claude-research (Claude-3.7-Sonnet) & 45.00 & 45.34 & 42.79 & 47.58 & 44.66 & - & - & 175 \\
        kimi-researcher (Kimi-K2) & 44.64 & 44.96 & 41.97 & 47.14 & 45.59 & - & - & 1000 \\
        doubao-deepresearch (Doubao-1.5-Pro) & 44.34 & 44.84 & 40.56 & 47.95 & 44.69 & 52.86 & 52.62 & 300 \\
        langchain-open-deep-research (GPT-4o) & 43.44 & 42.97 & 39.17 & 48.09 & 45.22 & 49.10 & 29.49 & 200 \\
        ours (\texttt{gpt-oss-20b}) & \textbf{41.45} & \textbf{40.22} & \textbf{37.39} & \textbf{45.72} & \textbf{44.87} & - & - & 20 \\
        nvidia-aiq-research-assistant & 40.52 & 37.98 & 38.39 & 44.59 & 42.63 & - & - & 70 \\
        tongyi-deepresearch-30B-A3B & 40.46 & 39.46 & 34.44 & 46.22 & 44.27 & - & - & 30 \\
        perplexity-Research (GPT-5) & 40.46 & 39.10 & 35.65 & 46.11 & 43.08 & 82.63 & 31.20 & 1000 \\
        grok-deeper-search (Grok-2) & 38.22 & 36.08 & 30.89 & 46.59 & 42.17 & 73.08 & 8.58 & 270 \\
        sonar-reasoning-pro & 37.76 & 34.96 & 31.65 & 44.93 & 42.42 & 45.19 & 9.39 & 670 \\
        sonar-reasoning & 37.75 & 34.73 & 32.59 & 44.42 & 42.39 & 52.58 & 13.37 & 70 \\
        claude-3-7-sonnet-with-search & 36.63 & 35.95 & 31.29 & 44.05 & 36.07 & 87.32 & 24.51 & - \\
        sonar-pro & 36.19 & 33.92 & 29.69 & 43.39 & 41.07 & 79.72 & 16.75 & 70 \\
        gemini-2.5-pro-preview & 31.90 & 31.75 & 24.61 & 40.24 & 32.76 & - & - & 128 \\
        gpt-4o-search-preview & 30.74 & 27.81 & 20.44 & 41.01 & 37.60 & 86.63 & 5.05 & 200 \\
        sonar & 30.64 & 27.14 & 21.62 & 40.70 & 37.46 & 76.41 & 10.68 & 20 \\
        gpt-4.1 & 29.31 & 25.59 & 18.42 & 40.63 & 36.49 & 89.85 & 4.27 & 200 \\
        gemini-2.5-flash-preview & 29.19 & 28.97 & 21.62 & 37.80 & 29.97 & - & - & 128 \\
        gpt-4o-mini-search-preview & 27.62 & 24.24 & 16.62 & 38.59 & 35.27 & 81.69 & 4.62 & 200 \\
        gpt-4.1-mini & 26.62 & 22.86 & 15.39 & 38.18 & 34.49 & 84.54 & 4.10 & 200 \\
        claude-3-5-sonnet-with-search & 23.95 & 21.28 & 16.20 & 32.41 & 29.87 & 94.06 & 9.35 & 175 \\
        \bottomrule
    \end{tabular}
    }
    \caption{Detailed performance breakdown on the DeepResearch benchmark. Scores for comprehensiveness (Comp.), insight, instruction following (Inst. Fol.), and readability (Read.) are normalized. Parameters are estimated based on public reports.}
    \label{tab:detailed_results}
\end{table*}

\end{document}